\documentclass[runningheads]{llncs}
\usepackage{graphicx}

\usepackage{tikz}
\usepackage{comment}
\usepackage{color}
\usepackage{multirow}

\usepackage[accsupp]{axessibility}  

\begin{document}
\sloppy
\pagestyle{headings}
\mainmatter
\def\ECCVSubNumber{3985}  

\title{Self-Constrained Inference Optimization on  Structural Groups  for Human Pose Estimation} 

\titlerunning{SCIO on  Structural Groups  for Human Pose Estimation}
%
\author{Zhehan Kan\and
Shuoshuo Chen\and
Zeng Li
\and
Zhihai He\thanks{Corresponding author}}
\authorrunning{Z. Kan et al.}
%
\institute{Southern University of Science and Technology\\
\email{\{kanzh2021, chenss2021\}@mail.sustech.edu.cn} \\
\email{\{liz9, hezh\}@sustech.edu.cn}
}
\maketitle

\begin{abstract}
We observe that human poses exhibit strong group-wise structural correlation and spatial coupling between keypoints due to the biological constraints of different body parts. 
This group-wise structural correlation can be explored to improve the accuracy and robustness of human pose estimation. 
In this work, we develop a self-constrained  prediction-verification network to characterize and learn the structural correlation between keypoints during training. During the inference stage, the feedback information from the verification network allows us to perform further optimization of pose prediction, which significantly improves the performance of human pose estimation.
Specifically, we partition the keypoints into groups according to the biological structure of human body. Within each group, the keypoints are further partitioned into two subsets, high-confidence base keypoints and low-confidence terminal keypoints. We develop a self-constrained prediction-verification network  to perform forward and backward predictions between these keypoint subsets. 
One fundamental challenge in pose estimation, as well as in generic prediction tasks, is that there is no mechanism for us to verify if the obtained pose estimation or prediction results are accurate or not, since the ground truth is not available. 
Once successfully learned, the verification network serves as an accuracy  verification module for the forward pose prediction. During the inference stage, it can  be used to guide the local optimization of the pose estimation results of low-confidence keypoints with the self-constrained loss on high-confidence keypoints as the objective function. Our extensive experimental results on benchmark MS COCO and CrowdPose datasets demonstrate that the proposed method can significantly improve the pose estimation results. 
\end{abstract}

\section{Introduction}
\label{sec:intro}
Human pose estimation aims to correctly detect and localize keypoints, i.e., human body joints or parts, for all persons in an input image. It is one of the fundamental computer vision tasks which plays an important role  in a variety of downstream applications, such as motion capture \cite{DBLP:conf/cvpr/ElhayekAJTPABST15,DBLP:conf/cvpr/RhodinCKSF19}, activity recognition \cite{DBLP:conf/cvpr/BagautdinovAFFS17,DBLP:conf/cvpr/WuWWGW19}, and person tracking \cite{DBLP:conf/cvpr/YangRLZW021,DBLP:conf/cvpr/WangTM20}. Recently, remarkable process has been made in human pose estimation based on deep neural network methods \cite{DBLP:conf/cvpr/CaoSWS17,Chen_2018_CVPR,DBLP:conf/cvpr/0009XLW19,He_2017_ICCV,DBLP:conf/cvpr/PapandreouZKTTB17,DBLP:conf/cvpr/SuYXGW19}. For regular scenes, deep learning-based methods have already achieved remarkably accurate estimation of body keypoints and there is little space for further performance improvement \cite{DBLP:conf/cvpr/ZhangZD0Z20,DBLP:conf/eccv/WangLGDW20,DBLP:conf/cvpr/0005ZGH20}. However,  for complex scenes with person-person occlusions, large variations of appearance, and cluttered backgrounds, pose estimation remains very challenging \cite{DBLP:conf/eccv/XiaoWW18,DBLP:conf/cvpr/0005ZGH20}. 
We  notice that, in complex scenes, the performance of pose estimation on different keypoints exhibits large variations. For example, for those visible keypoints with little interference from other persons or background, their estimation results are fairly accurate and reliable. However, for some keypoints, for example the terminal keypoints at tip locations of body parts, it is very challenging to achieve accurate pose estimation. The low accuracy of these challenging keypoints degrades the overall pose estimation performance. Therefore, the main challenge in pose estimation is how to improve the estimation accuracy of these challenging keypoints.

\begin{figure}[t]
\centering
\setlength{\belowcaptionskip}{-0.4cm} 
\includegraphics[width=0.6\columnwidth]{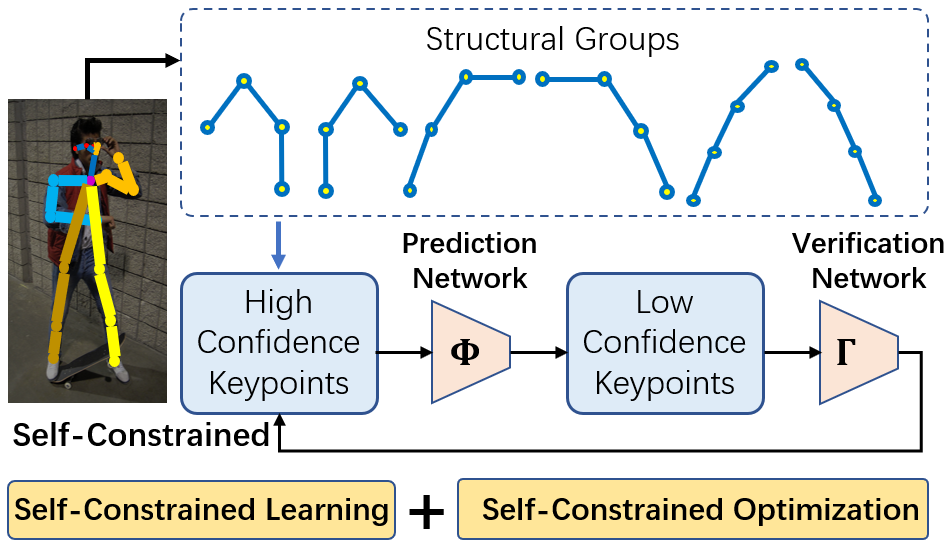}
\centering
\caption{Illustration of the proposed idea of self-constrained inference optimization of structural groups for human pose estimation.}
\label{fig:idea}
\end{figure}

 As summarized in Fig. \ref{fig:idea}, this work is motivated by the following two important observations: (1) human poses, although exhibiting large variations due to the free styles and flexible movements of human, are however restricted by the biological structure of the body. The whole body consists of multiple  parts, such as the upper limbs and lower limbs. Each body part corresponds to a subgroup of keypoints. We observe that the keypoint correlation across different body parts remains low since different body parts, such as the left and right arms, can move with totally different styles and towards different directions. However, within the same body part or within the same structural group, keypoints are more spatially constrained by each other. This implies that keypoints can be potentially predictable from each other by exploring this unique structural correlation. Motivated by this observation, in this work, we propose to partition the body parts into a set of structural groups and perform group-wise structure learning and keypoint prediction refinement. 

\begin{figure}[h]
\centering
\setlength{\abovecaptionskip}{-0.1cm} 
\setlength{\belowcaptionskip}{-0.4cm} 
\includegraphics[width=0.95\columnwidth]{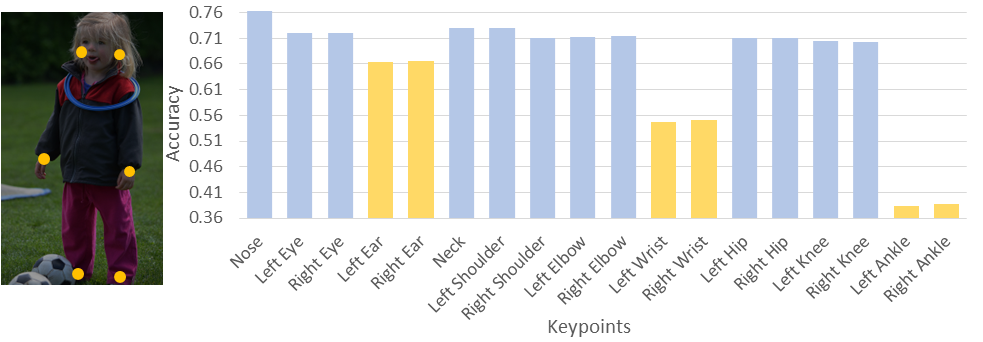}
\centering
\caption{Keypoints at the tip locations of body parts suffer from low confidence scores obtained from the heatmap during pose estimation.}
\label{fig:confidence}
\end{figure}

(2) We have also observed that, within each group of keypoints, terminal keypoints at tip locations of the body parts, such as 
ankle and wrist keypoints, often suffer from lower estimation accuracy. This is because they have much larger freedom of motion and are more easily to be occluded by other objects. Fig. \ref{fig:confidence} shows the average prediction confidence (obtained from the heatmaps) of all keypoints with yellow dots and bars representing the locations and estimation confidence for terminal keypoints, for example, wrist or ankle keypoints.  We can see that the average estimation confidence of terminal keypoints are much lower than the rest. 

Motivated by the above two observations, we propose to partition the body keypoints into 6 structural groups according to their biological parts, and each structural group is further partitioned into two subsets: \textit{terminal keypoints} and \textit{base keypoints} (the rest keypoints). We develop a self-constrained prediction-verification network to learn the structural correlation between these two subsets within each structural group. Specifically, we learn two tightly coupled networks, the prediction network $\mathbf{\Phi}$ which performs the forward prediction of terminal keypoints from base keypoints, and the verification network $\mathbf{\Gamma}$ which performs backward prediction of the base keypoints from terminal keypoints. 
This prediction-verification network aims to characterize the structural correlation between keypoints within each structural group. They are jointly learned using a self-constraint loss.
Once successfully learned, the verification network $\mathbf{\Gamma}$ is then used as a performance assessment module to optimize the prediction of low-confidence terminal keypoints based on local search and refinement within each structural group.
Our extensive experimental results on benchmark MS COCO datasets demonstrate that the proposed method is able to significantly improve the pose estimation results. 

The rest of the paper is organized as follows. Section 2 reviews related work on human pose estimation. The proposed 
self-constrained inference optimization of structural groups is presented in Section 3. Section 4 presents the experimental results, performance comparisons, and ablation studies. Section 5  concludes the paper.

\section{Related Work and Major Contributions}
\label{sec:related}
In this section, we review related works on heatmap-based pose estimation, multi-person pose estimation, pose refinement and error correction, and reciprocal learning. We then summarize the major contributions of this work.

\textbf{(1) Heatmap-based pose estimation.}
In this paper, we use heatmap-based pose estimation. 
The probability for a pixel to be the keypoint can be measured by its response in the heatmap. Recently, heatmap-based approaches have achieved the state-of-the-art performance in pose estimation \cite{DBLP:conf/eccv/XiaoWW18,Cheng_2020_CVPR,DBLP:conf/cvpr/XuT21,DBLP:conf/cvpr/0009XLW19}.  The  coordinates of keypoints are obtained by decoding the heatmaps \cite{DBLP:conf/cvpr/SuYXGW19}.  \cite{Cheng_2020_CVPR} predicted scale-aware high-resolution heatmaps using multi-resolution aggregation during inference. \cite{DBLP:conf/cvpr/XuT21} processed graph-structured features across multi-scale human skeletal representations and proposed a learning approach for multi-level feature learning and heatmap estimation.

\textbf{(2) Multi-person pose estimation.}
Multi-person pose estimation requires detecting keypoints of all persons in an image \cite{Fang_2017_ICCV}. It is very challenging due to overlapping between body parts from neighboring persons. Top-down methods and bottom-up methods have been developed in the literature to address this issue. \textbf{(a) Top-down} approaches \cite{He_2017_ICCV,DBLP:conf/eccv/SunXWLW18,DBLP:conf/cvpr/MoonCL19,DBLP:conf/cvpr/SuYXGW19} first detect all persons in the image and then estimates keypoints of each person. The performance of this method depends on the reliability of object detection which generates the bounding box for each person. When the number of persons is large, accurate detection of each person becomes very challenging, especially in highly occluded and cluttered scenes \cite{DBLP:conf/cvpr/PapandreouZKTTB17}.
\textbf{(b) Bottom-up} approaches \cite{Geng_2021_CVPR,DBLP:conf/cvpr/CaoSWS17,Luo_2021_CVPR} directly detect keypoints of all persons and then group keypoints for each person. These methods usually run faster than the top-down methods in multi-person pose estimation since they do not require person detection. \cite{Geng_2021_CVPR} activated the pixels in the keypoint regions and learned disentangled representations for each keypoint to improve the regression result. \cite{Luo_2021_CVPR} developed a scale-adaptive heatmap regression method to handle large variations of body sizes.

\textbf{(3) Pose refinement and error correction.} A number of methods have been developed in the literature to refine the estimation of body keypoints \cite{9107502,DBLP:conf/cvpr/MoonCL19,DBLP:conf/eccv/WangLGDW20}.
\cite{8575519} proposed a pose refinement network  which takes  the image and the predicted keypoint locations as input and learns to directly predict refined keypoint locations. \cite{9107502} designed two networks where the correction network guides the refinement to correct the joint locations before generating the final pose estimation. \cite{DBLP:conf/cvpr/MoonCL19} introduced a model-agnostic pose refinement method using statistics of error distributions as prior information to generate synthetic poses for training.  \cite{DBLP:conf/eccv/WangLGDW20}  introduced a localization sub-net to extract different visual features and a graph pose refinement module to explore the relationship between points sampled from the heatmap regression network. 

\textbf{(4) Cycle consistency and reciprocal learning.}
This work is related to cycle consistency and reciprocal learning. \cite{Zhu_2017_ICCV} translated an image from the source domain into the target domain by introducing a cycle consistence constraint so that the distribution of images from translated domain is indistinguishable from the distribution of target domain. 
\cite{Sun_2020_CVPR} developed a pair of jointly-learned networks to predict human trajectory forward and backward. 
\cite{xu2020segmentation} developed a reciprocal cross-task architecture for image segmentation, which improves the learning efficiency and generation accuracy by exploiting the commonalities and differences across tasks. 
\cite{liu2021watching} developed a  Temporal Reciprocal Learning (TRL) approach to fully explore the discriminative information from the disentangled features. \cite{zhang2021accurate} designed a support-query mutual guidance architecture for few-shot object detection.

\vspace{0.2cm}
\textbf{(5) Major contributions of this work.} 
Compared to the above related work, the major contributions of this work are: (a) we propose to partition the body keypoints into structural groups and explore the structural correlation within each group to improve the pose estimation results. 
Within each structural group, we propose to partition the keypoints into high-confidence and low-confidence ones. We  develop a  prediction-verification network to characterize 
structural correlation between them based on a self-constraint loss. (b) We introduce a self-constrained  optimization method which uses the learned verification network as a performance assessment module to optimize the pose estimation of low-confidence keypoints during the inference stage. (c) Our extensive experimental results have demonstrated that our proposed method is able to significantly improve the performance of pose estimation and outperforms the existing methods by large margins. 

Compared to existing methods on cycle consistency and reciprocal learning, our  method has the following unique novelty. First, it addresses an important problem in prediction: how do we know if the prediction is accurate or not since we do not have the ground-truth. It establishes a self-matching constraint on high-confidence keypoints and uses the successfully learned verification network to verify if the refined predictions of low-confidence keypoints are accurate or not. Unlike existing prediction methods which can only perform forward inference, our method 
is able to perform further optimization of the prediction results during the inference stage, which can significantly improve the prediction accuracy and the generalization capability of the proposed method.


\section{Method}
In this section, we present our self-constrained inference optimization  (SCIO) of structural groups for human pose estimation.

\subsection{Problem Formulation}
Human pose estimation, as a keypoint detection task, aims to detect the locations of body keypoints from the input image. Specifically, let $I$ be the image of size $W \times H \times 3$. Our task is to locate $K$ keypoints $X=\{X_1,X_2, ...,X_K\}$ from $I$ precisely. Heatmap-based methods transform this problem to estimate $K$ heatmaps $\{H_1,H_2, ...,H_K\}$ of size $W’ \times H’$. Given a heatmap, the keypoint location can be determined using different grouping or peak finding methods \cite{DBLP:conf/cvpr/MoonCL19,DBLP:conf/cvpr/SuYXGW19}. 
For example, the pixel with the highest heatmap value can be designated as the location of the corresponding keypoint. 
Meanwhile, given a keypoint at location $(p_x, p_y)$, the corresponding heatmap can be generated using the Gaussian kernel

\begin{equation}
    C(x, y) = \frac{1}{2\pi \sigma^2} e^{-[(x-p_x)^2 + (y-p_y)^2]/2\sigma^2}.
\end{equation}
In this work, the ground-truth heatmaps are denoted by 
${\bar{H}_1,\bar{H}_2, ..., \bar{H}_K}$.

\begin{figure*}[t]
\centering
\setlength{\abovecaptionskip}{-0.05cm} 
\setlength{\belowcaptionskip}{-0.8cm} 
\includegraphics[width=0.99\columnwidth]{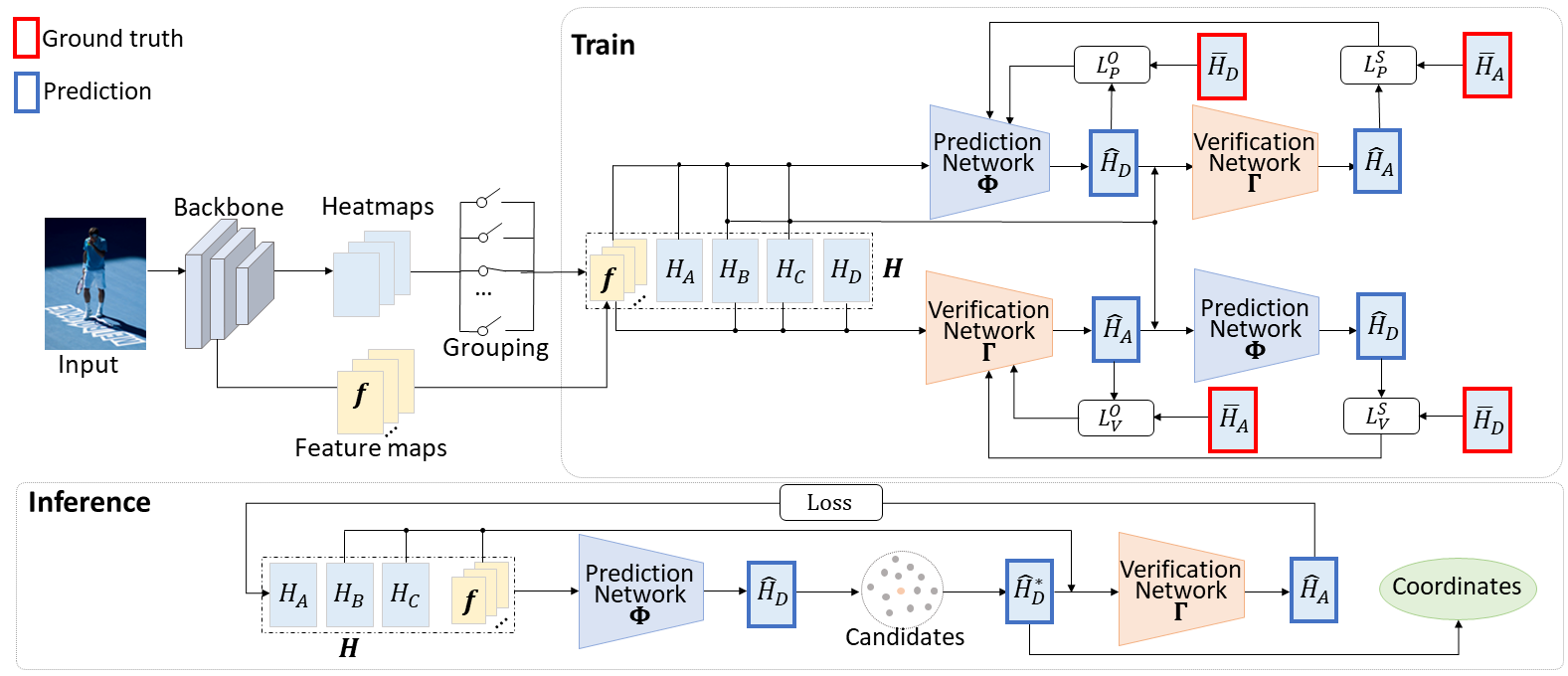}
\centering
\caption{The overall framework of our proposed network. For an input image, heatmaps of all keypoints predicted by the  backbone are partitioned into 6 structural groups. During training stage, each group $\mathbf{H}$ is divided into two subsets: base keypoints and terminal keypoints. A prediction-verification network with self-constraints is developed to characterize the structural correlation between these two subsets. During testing, the learned verification network is used to refine the prediction results of the low-confidence terminal keypoints. }
\label{fig:framework}
\end{figure*}

\subsection{Self-Constrained Inference Optimization on Structural Groups}
\label{sec:overview}
\begin{figure}[h]
\centering
\setlength{\abovecaptionskip}{-0.05cm} 
\includegraphics[width=0.5\columnwidth]{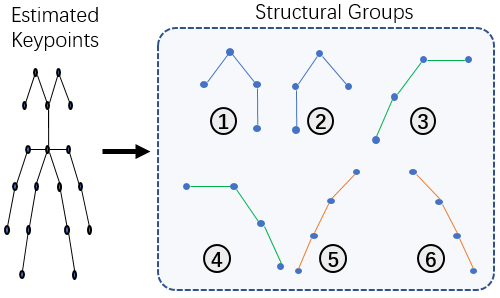}
\centering
\caption{Partition of the body keypoints into 6 structural groups corresponding to different body parts. Each group has 4 keypoints.}
\label{fig:groups}
\end{figure}

Fig. \ref{fig:framework} shows the overall framework of our proposed SCIO method for pose estimation. 
We first partition the detected human body keypoints into 6 structural groups, which correspond to different body parts, including lower and upper limbs, as well as two groups for the head part, as illustrated in Fig. \ref{fig:groups}.
Each group contains four keypoints. We observe that these structural groups of four keypoints are the basic units for human pose and body motion. They are constrained by the biological structure of the human body.  There are significant freedom and variations between structural groups. For example, the left arm and the right arm could move and pose in totally different ways. In the meantime, within each group, the set of keypoints are constraining each other with strong structural correlation between them. 

As discussed in Section \ref{sec:intro}, we further partition each of these 6 structural groups into base keypoints and terminal keypoints. The base keypoints are near the body torso while the terminal keypoints are at the end or tip locations of the corresponding body part. 
Fig. \ref{fig:confidence} shows that the terminal keypoints are having much lower estimation confidence scores than those base keypoints during pose estimation. 
In this work, we denote these 4 keypoints within each group by  
\begin{equation}
    \mathbf{G} = \{X_A, X_B, X_C \ |\  X_D\},
\end{equation} 
where $X_D$ is the terminal keypoint and the rest three $\{X_A, X_B, X_C\}$ are the base keypoints near the torso. 
The corresponding heatmap are denoted by 
$\mathbf{H} = \{H_A, H_B, H_C \ |\  H_D\}$. 
To characterize the structural correlation within each structural group  $\mathbf{H}$, we propose to develop a self-constrained prediction-verification network. As illustrated in Fig. \ref{fig:framework}, the prediction network $\mathbf{\Phi}$ predicts the 
heatmap of the terminal keypoint $H_D$ from the base keypoints 
$\{H_A, H_B, H_C\}$ with  feature map $\mathbf{f}$ as the visual context:
\begin{equation}
    \hat{H}_D = \mathbf{\Phi}(H_A, H_B, H_C; \mathbf{f}).
    \label{eq:prediction}
\end{equation}
We observe that the feature map $\mathbf{f}$  provides important visual context for keypoint estimation.
The verification network $\mathbf{\Gamma}$ shares the same structure as the prediction network. It performs the backward prediction of keypoint $H_A$ from the rest three:
\begin{equation}
    \hat{H}_A = \mathbf{\Gamma}(H_B, H_C, H_D; \mathbf{f}).
    \label{eq:verifiction}
\end{equation}
Coupling the prediction and verification network together by passing the 
prediction output $\hat{H}_D$ of the prediction network into the verification network as input, we have the following prediction loop
\begin{eqnarray}
   \hat{H}_A &=& \mathbf{\Gamma}(H_B, H_C, \hat{H}_D; \mathbf{f})\\
   &=& \mathbf{\Gamma}(H_B, H_C, \mathbf{\Phi}(H_A, H_B, H_C; \mathbf{f}); \mathbf{f}).
\end{eqnarray}
This leads to the following self-constraint loss
\begin{equation}
    \mathcal{L}_A^s = ||\bar{H}_A - \hat{H}_A||_2.
    \label{eq:selfloss}
\end{equation}
This  prediction-verification network with a forward-backward prediction loop
learns the internal structural correlation between the base keypoints and the terminal keypoint. The learning process is guided by the self-constraint loss. If the internal structural correlation is successfully learned, then the self-constraint loss 
$\mathcal{L}_A^s$ generated by the forward and backward prediction loop should be small. 
This step is referred to as \textit{self-constrained learning}.

Once successfully learned, the verification network $\mathbf{\Gamma}$ can be used to verify if the prediction $\hat{X}_D$ is accurate or not. 
In this case, the self-constraint loss is used as an objective function to optimize the prediction $\hat{X}_D$ based on local search, which can be formulated as 
\begin{eqnarray}
    \hat{X}_D^* &=& \arg\min_{\hat{X}_D}  ||H_A - \hat{H}_A||_2, \\
    &=& \arg\min_{\hat{X}_D} ||H_A - \mathbf{\Gamma}(H_B, H_C, \mathbb{H}(\hat{X}_D); \mathbf{f})||_2 \nonumber,
\end{eqnarray}
where $\mathbb{H}(\hat{X}_D)$ represents the heatmap generated from keypoint $\hat{X}_D$ using the Gaussian kernel.
This provides an effective mechanism for us to iteratively refine the prediction result based on the specific statistics of  the test sample. This adaptive prediction and optimization is not available in traditional network prediction which is purely forward without any feedback or adaptation. This feedback-based adaptive prediction will result in better generalization capability on the test sample. 
This step is referred to as \textit{self-constrained optimization}.
In the following sections, we present more details about the proposed self-constrained learning (SCL) and self-constrained optimization (SCO) methods.

\subsection{Self-Constrained Learning of Structural Groups}
In this section, we explain the self-constrained learning in more details. 
As illustrated in Fig. \ref{fig:framework}, the input to the prediction and verification networks, namely,  $\{H_A, H_B, H_C\}$ and $\{H_B, H_C, H_D\}$, are all heatmaps generated by the baseline pose estimation network. In this work, we use the HRNet \cite{DBLP:conf/cvpr/0009XLW19} as our baseline, on top of which our proposed SCIO method is implemented. We observe that the visual context surrounding the keypoint location provides important visual cues for  refining the locations of the keypoints. For example, the correct location of the knee keypoint should be at the center of the knee image region. Motivated by this, we also pass the feature map $\mathbf{f}$ generated by the backbone network to the prediction and verification network as inputs. 

In our proposed scheme of self-constrained learning, the prediction and verification networks are jointly trained. Specifically, as illustrated in Fig. \ref{fig:framework}, the top branch shows the training process of the prediction network. Its input includes heatmaps 
$\{H_A, H_B, H_C\}$  and the visual feature map $\mathbf{f}$.
The output of the prediction network is the predicted heatmap for keypoint $X_D$, denoted by $\hat{H}_D$. 
During the training stage, this prediction is compared to its ground-truth $\bar{H}_D$ and form the prediction loss $\mathcal{L}_P^O$ which is given by 
\begin{equation}
    \mathcal{L}_P^O = ||\hat{H}_D - \bar{H}_D||_2.
\end{equation}
The predicted heatmap $\hat{H}_D$, combined with 
the heatmaps $H_B$ and $H_C$ and the visual feature map $\mathbf{f}$, is passed to the verification network $\mathbf{\Gamma}$ as input.  The output of $\mathbf{\Gamma}$ will be the predicted heatmap for keypoint $X_A$, denoted by $\hat{H}_A$.  
We then compare it with the ground-truth heatmap $\bar{H}_A$ and define the following self-constraint loss for the prediction network
\begin{equation}
     \mathcal{L}_P^S = ||\hat{H}_A - \bar{H}_A||_2.
\end{equation}
These two losses are combined as $\mathcal{L}_P = \mathcal{L}_P^O+\mathcal{L}_P^S$ to train the prediction network $\mathbf{\Phi}$.

Similarly, for the verification network, the inputs are heatmaps 
$\{H_B, H_C, H_D\}$ and visual feature map $\mathbf{f}$. It predicts the heatmap $\hat{H}_A$ for keypoint $X_A$. It is then, combined with $\{H_B, H_C\}$ and $\mathbf{f}$ to form the input to the prediction network $\mathbf{\Phi}$ which predicts the 
heatmap $\hat{H}_D$. Therefore, the overall loss function for the verification network is given by 
\begin{equation}
     \mathcal{L}_V = ||\hat{H}_A - \bar{H}_A||_2 +
     ||\hat{H}_D - \bar{H}_D||_2.
\end{equation}
The prediction and verification network are jointly trained in an iterative manner. Specifically, during the training epochs for the prediction network, the verification network is fixed and used to compute the self-constraint loss for the prediction network. Similarly, during the training epochs for the verification network, the prediction network is fixed and used to compute the self-constraint loss for the verification network.

\subsection{Self-Constrained Inference Optimization of Low-Confidence Keypoints}
\label{sec:slo}

As discussed in Section \ref{sec:intro}, one of the major challenges in pose estimation is to improve the accuracy of hard keypoints, for example, those terminal keypoints. 
In existing approaches for network prediction, the inference process is purely forward. The knowledge learned from the training set is directly applied to the test set. There is no effective mechanism to verify if the prediction result is accurate or not since the ground-truth is not available. This  forward inference process often suffers from generalization problems since  there is no feedback process to adjust the prediction results based on the actual test samples. 

The proposed self-constrained inference optimization aims to address the above issue. The verification network $\mathbf{\Gamma}$, once successfully learned, can be used as a feedback module to evaluate the accuracy of the prediction result. This is achieved by mapping the prediction result $\hat{H}_D$ for the low-confidence keypoint back to the high-confidence keypoint $\hat{H}_A$. Using the self-constraint loss as an objective function, we can perform local search or refinement of the prediction result $\hat{X}_D$ to minimize the objective function, as formulated in (8). 
Here, the basic idea is that: if the prediction $\hat{X}_D$ becomes accurate during local search, then, using it as the input, the verification network should be able to accurately predict the high-confidence keypoint $\hat{H}_A$, which implies that the 
self-constraint loss $||{H}_A - \hat{H}_A||_2$ on the high-confidence keypoint ${X}_A$ should be small.   

Motivated by this, we propose to perform local search and refinement of the low-confidence keypoint. Specifically, we add a small perturbation $\Delta_D$ onto the predicted result 
$\hat{X}_D$ and search its small neighborhood to minimize the self-constraint loss:
\begin{equation}
    \hat{X}_D^{*} = \arg\min_{\tilde{H}_D } ||H_A - \mathbf{\Gamma}(H_B, H_C, \tilde{H}_D; \mathbf{f})||_2 \nonumber
\end{equation}
\begin{equation}
    \tilde{H}_D = \mathbb{H}(\hat{X}_D +\Delta_D),\ \  ||\Delta_D||_2 \le \delta.
\end{equation}
Here, $\delta$ controls the search range and direction of the keypoint, and the direction will be dynamically adjusted with the loss.
$\mathbb{H}(\hat{X}_D +\Delta_D)$ represents the heatmap generated from the keypoint location $\hat{X}_D +\Delta_D$ using the Gaussian kernel.
In the Supplemental Material section, we provide further discussion on the extra computational complexity of the proposed SCIO method. 

\begin{table}
\begin{center}
\caption{
Comparison with the state-of-the-arts methods on COCO test-dev.
}
\label{tab:sota on COCO}
\begin{tabular}{l|l|ccccccc}
\hline\noalign{\smallskip}
Method & Backbone & Size & \text{$AP$} & $AP^{50}$ & $AP^{75}$ &  $AP^{M}$ & $AP^{L}$ & $AR$  \\
\hline\noalign{\smallskip}
CMU-Pose \cite{DBLP:conf/cvpr/CaoSWS17} & - & - & 61.8 & 84.9 & 67.5 & 57.1 & 68.2 & 66.5\\
Mask-RCNN \cite{He_2017_ICCV} & R50-FPN & - & 63.1 & 87.3 & 68.7 & 57.8 & 71.4 & - \\
G-RMI \cite{DBLP:conf/cvpr/PapandreouZKTTB17} & R101 & 353$\times$257 & 64.9& 85.5 &71.3& 62.3&70.0 &69.7\\
AE \cite{DBLP:conf/nips/NewellHD17} & - &  512$\times$512& 65.5& 86.8& 72.3& 60.6& 72.6 &70.2\\
Integral Pose \cite{DBLP:conf/eccv/SunXWLW18} & R101 & 256$\times$256& 67.8 &88.2& 74.8 &63.9 &74.0 &-\\
RMPE \cite{Fang_2017_ICCV} &PyraNet& 320$\times$256& 72.3& 89.2 &79.1& 68.0& 78.6& -\\
CFN \cite{DBLP:conf/iccv/HuangGT17} & -& -& 72.6& 86.1& 69.7& \textbf{78.3}& 64.1& -\\
CPN(ensemble) \cite{Chen_2018_CVPR}& ResNet-Incep. &384$\times$288 &73.0& 91.7& 80.9 &69.5& 78.1 &79.0\\
CSM+SCARB \cite{DBLP:conf/cvpr/SuYXGW19} & R152& 384$\times$288 &74.3 &91.8& 81.9 &70.7 &80.2 &80.5\\
CSANet \cite{DBLP:journals/corr/abs-1905-05355} & R152& 384$\times$288& 74.5& 91.7 &82.1& 71.2 &80.2& 80.7\\
HRNet \cite{DBLP:conf/cvpr/0009XLW19} & HR48& 384$\times$288 &75.5& 92.5& 83.3& 71.9 &81.5& 80.5\\
MSPN \cite{DBLP:journals/corr/abs-1901-00148} & MSPN &384$\times$288 &76.1& {93.4} &83.8& 72.3 &81.5& 81.6\\
DARK \cite{DBLP:conf/cvpr/ZhangZD0Z20} &  HR48& 384$\times$288 &76.2& 92.5& 83.6& 72.5 &82.4& 81.1\\
UDP \cite{DBLP:conf/cvpr/0005ZGH20} &  HR48 &384$\times$288& 76.5 &92.7& 84.0& 73.0& 82.4& 81.6\\
PoseFix \cite{DBLP:conf/cvpr/MoonCL19}&  HR48+R152& 384$\times$288 &76.7 &92.6& 84.1& 73.1& 82.6& 81.5\\
Graph-PCNN \cite{DBLP:conf/eccv/WangLGDW20} & HR48 &384$\times$288 &76.8& 92.6& 84.3& 73.3& 82.7 &81.6\\
\hline\noalign{\smallskip}
\textbf{SCIO} (Ours) & HR48 & 384$\times$288 & \textbf{79.2} &  \textbf{93.5} & \textbf{85.8} & 74.1 & \textbf{84.2} & \textbf{81.6}\\
\textbf{Performance Gain} & & & \textbf{+2.4} &\textbf{+0.9}&\textbf{+1.5}&&\textbf{+1.5}&\textbf{+0.0}\\
\hline\noalign{\smallskip}
\end{tabular}
\end{center}
\end{table}
\vspace{-1.0cm}

\section{Experiments}
In this section, we present  experimental results, performance comparisons with state-of-the-art methods, and ablation studies to demonstrate the performance of our SCIO method.

\subsection{Datasets}
The comparison and ablation experiments are performed on MS COCO dataset \cite{DBLP:conf/eccv/LinMBHPRDZ14} and CrowdPose \cite{DBLP:conf/cvpr/LiWZMFL19} dataset, both of which contain very challenging scenes for pose estimation.

\textbf{MS COCO Dataset}: The COCO dataset contains challenging images with multi-person poses of various body scales and occlusion patterns in unconstrained environments. It contains 64K images and 270K persons labeled with 17 keypoints. We train our models on train2017 with 57K images including 150K persons and conduct ablation studies on val2017. We test our models on test-dev for performance comparisons with the state-of-the-art methods. In evaluation, we use the metric of Object Keypoint Similarity (OKS) score to evaluate the performance.

\textbf{CrowdPose Dataset}: The CrowdPose dataset contains 20K images and 80K persons labeled with 14 keypoints. Note that, for this dataset, we partition the keypoints into 4 groups, instead of 6 groups as in the COCO dataset. CrowdPose has more crowded scenes.
For training, we use the train set which has 10K images and 35.4K persons. For evaluation, we use the validation set which has 2K images and 8K persons, and the test set which has 8K images and 29K persons. 

\begin{table}[t]
\begin{center}
\caption{
Comparison with the state-of-the-arts methods on CrowdPose test-dev.
}
\label{tab:sota on Crowdpose}
\begin{tabular}{l|cccccccc}
\hline\noalign{\smallskip}
Method & Backbone & \text{$AP$}  & $AP^{med}$  \\
\hline\noalign{\smallskip}
Mask-RCNN \cite{He_2017_ICCV} & ResNet101 & 60.3  & -   \\
OccNet \cite{DBLP:conf/avss/GoldaKSB19} & ResNet50 & 65.5  & 66.6 \\
JC-SPPE \cite{DBLP:conf/cvpr/LiWZMFL19} & ResNet101 & 66  & 66.3    \\
HigherHRNet \cite{Cheng_2020_CVPR} &HR48 & 67.6  & -  \\
MIPNet \cite{Khirodkar_2021_ICCV} &HR48 & 70.0  & 71.1  \\
\hline\noalign{\smallskip}
\textbf{SCIO} (Ours) & HR48  & \textbf{71.5}  & \textbf{72.2} \\
\textbf{Performance Gain} &   & \textbf{+1.5}   & \textbf{+1.1}   \\
\hline\noalign{\smallskip}
\end{tabular}
\end{center}
\end{table}

\begin{table}[t]
\begin{center}
\caption{
Comparison with state-of-the-art of three backbones on COCO test-dev.
}
\label{tab:backbone}
\begin{tabular}{l|cccccccc}
\hline\noalign{\smallskip}
Method & Backbone & Size & \text{$AP$} & $AP^{50}$ & $AP^{75}$ &  $AP^{M}$ & $AP^{L}$ & $AR$  \\
\hline\noalign{\smallskip}
SimpleBaseline \cite{DBLP:conf/eccv/XiaoWW18} & R152 &384$\times$288& 73.7& 91.9& 81.1 &70.3& 80.0 &79.0\\
SimpleBaseline & \multirow{2}{*}{R152} & \multirow{2}{*}{384$\times$288} & \multirow{2}{*}{\textbf{77.9}} & \multirow{2}{*}{\textbf{92.1}} & \multirow{2}{*}{\textbf{82.7}} & \multirow{2}{*}{\textbf{72.6}} & \multirow{2}{*}{\textbf{82.3}} & \multirow{2}{*}{\textbf{80.9}}\\
+\textbf{SCIO} (Ours)\\
\textbf{Performance Gain} &   &   & \textbf{+4.2} & \textbf{+0.2} & \textbf{+1.6} & \textbf{+2.3}& \textbf{+2.3} & \textbf{+1.9}\\
\hline\noalign{\smallskip}

HRNet \cite{DBLP:conf/cvpr/0009XLW19} & HR32& 384$\times$288& 74.9& 92.5& 82.8& 71.3& 80.9 &80.1\\
HRNet+\textbf{SCIO} (Ours) & HR32 & 384$\times$288 & \textbf{78.6} & \textbf{92.7} & \textbf{84.2} & \textbf{73.3} & \textbf{82.9} & \textbf{81.5}\\
\textbf{Performance Gain} &   &   & \textbf{+3.7} & \textbf{+0.2}  & \textbf{+1.4} & \textbf{+2.0}& \textbf{+2.0} & \textbf{+1.4}\\
\hline\noalign{\smallskip}

HRNet \cite{DBLP:conf/cvpr/0009XLW19} & HR48& 384$\times$288 &75.5& 92.5& 83.3& 71.9 &81.5& 80.5\\
HRNet+\textbf{SCIO} (Ours) & HR48 & 384$\times$288 & \textbf{79.2} &  \textbf{93.5} & \textbf{85.8} & 74.1 & \textbf{84.2} & \textbf{81.6}\\
\textbf{Performance Gain} & & & \textbf{+3.7} &\textbf{+1.0}&\textbf{+1.5}&\textbf{+2.2}&\textbf{+2.2}&\textbf{+0.0}\\
\hline\noalign{\smallskip}
\end{tabular}
\end{center}
\end{table}

\subsection{Implementation Details}
For fair comparisons, we use HRNet and ResNet as our backbone and follow the same training configuration as \cite{DBLP:conf/eccv/XiaoWW18} and \cite{DBLP:conf/cvpr/0009XLW19} for ResNet and HRNet, respectively. For the prediction and verification networks, we choose the FCN network \cite{long2015fully}.
The networks are  trained with the Adam optimizer. We choose a batch size of 36 and an initial learning rate of 0.001. The whole model is trained for 210 epochs. During inference, we set the number of search steps to be 50.

\subsection{Evaluation Metrics and Methods}
Following existing papers \cite{DBLP:conf/cvpr/0009XLW19}, we use the standard  Object Keypoint Similarity (OKS) metric which is defined as: 
\begin{equation}
OKS = \frac{\sum\limits_{i}e^{-d_i^2/2s^2k_i^2}\cdot \delta(v_i>0)}{\sum\limits_i\delta(v_i>0)}.
\end{equation}
Here $d_i$ is the Euclidean distance between the detected keypoint and the corresponding ground truth, $v_i$ is the visibility flag of the ground truth, $s$ is the object scale, and $k_i$ is a per-keypoint constant that controls falloff. $\delta(*)$ means if * holds, $\delta(*)$ equals to 1, otherwise, $\delta(*)$ equals to 0. We report standard average precision and recall scores:   $AP^{50}$, $AP^{75}$, $AP$, $AP^{M}$, $AP^{L}$, $AR$, $AP^{easy}$, $AP^{med}$, $AP^{hard}$ at various OKS \cite{Geng_2021_CVPR,DBLP:conf/cvpr/0009XLW19}.

\begin{table}
\begin{center}
\caption{
Comparison with DARK and Graph-GCNN of input size 128$\times$96 on COCO val2017.
}
\label{tab:inputsize}
\begin{tabular}{l|cccccccc}
\hline\noalign{\smallskip}
Method & Backbone & Size & \text{$AP$} & $AP^{50}$ & $AP^{75}$ &  $AP^{M}$ & $AP^{L}$ & $AR$  \\
\hline\noalign{\smallskip}
DARK \cite{DBLP:conf/cvpr/ZhangZD0Z20} &  HR48& 128$\times$96 &71.9&  89.1 & 79.6 & 69.2 & 78.0  &77.9\\
Graph-PCNN \cite{DBLP:conf/eccv/WangLGDW20}& HR48& 128$\times$96& 72.8& 89.2& 80.1& 69.9 &79.0 &78.6\\
\hline\noalign{\smallskip}
\textbf{SCIO} (Ours) & HR48& 128$\times$96& \textbf{73.7}& \textbf{89.6}& \textbf{80.9} & \textbf{70.3}& \textbf{79.4} & \textbf{79.1}\\
\textbf{Performance Gain} &   &   & \textbf{+0.9} & \textbf{+0.4}  & \textbf{+0.8} & \textbf{+0.4}& \textbf{+0.9} & \textbf{+0.8}\\
\hline\noalign{\smallskip}
\end{tabular}
\end{center}
\end{table}
\vspace{-1cm}

\subsection{Comparison to State of the Art}
We compare our SCIO method with other top-performing methods on the COCO test-dev and CrowdPose datasets. Table \ref{tab:sota on COCO} shows the performance comparisons with state-of-the-art methods on the MS COCO dataset. It should be noted that the best performance is reported here for each method. We can see that our SCIO method outperforms the current best by a large margin, up to 2.5\%, which is quite significant. 
Table \ref{tab:sota on Crowdpose} shows the results on challenging CrowdPose. In the literature, there are only few methods have reported results on this challenging dataset. Compared to the current best method MIPNet \cite{Khirodkar_2021_ICCV}, our SCIO method has improved the pose estimation accuracy by up to 1.5\%, which is quite significantly.

In Table \ref{tab:backbone}, we compare our SCIO with state-of-the-art methods using different backbone networks, including R152, HR32, and HR48 backbone networks. 
We can see that our SCIO method consistently outperforms existing methods. Table \ref{tab:inputsize} shows the performance comparison on pose estimation with different input image size, for example 128$\times$96 instead of 384$\times$288. We have only found two methods that reported results on small input images. 
We can see that our SCIO method also outperforms these two methods on small input images.

\begin{table}[t]
\begin{center}
\caption{Ablations study on COCO val2017.} 
\label{tab:ablations}
\begin{tabular}{l|cccccccc}
\hline\noalign{\smallskip}
  & $AP$ & $AP^{50}$ & $AP^{75}$    & $AR$ \\
\hline\noalign{\smallskip}
Baseline & 76.3  & 90.8  & 82.9    & 81.2\\
Baseline + SCL & 78.3 & 92.9 & 84.9  & 81.3\\
Baseline + SCL + SCO &\textbf{79.5} & \textbf{93.7} & \textbf{86.0} & \textbf{81.6} \\
\hline\noalign{\smallskip}
\end{tabular}
\end{center}
\end{table}

\begin{table}
\begin{center}
\caption{Ablations study of terminal keypoints accuracy on COCO val2017.}
\label{tab:ablations of keypoints}
\begin{tabular}{l|cccccc}
\hline\noalign{\smallskip}
 & Left & Right & Left & Right & Left & Right\\
 & Ear & Ear & Wrist & Wrist & Ankle & Ankle\\
\hline\noalign{\smallskip}
HRNet & 0.6637  & 0.6652  & 0.5476   & 0.5511 & 0.3843 & 0.3871\\
\hline\noalign{\smallskip}
HRNet + \textbf{SCIO}(Ours) &\textbf{0.7987} & \textbf{0.7949} & \textbf{0.7124} & \textbf{0.7147} & \textbf{0.5526} & \textbf{0.5484}\\
\textbf{Performance Gain} & \textbf{+0.1350} & \textbf{+0.1297}  & \textbf{+0.1648} & \textbf{+0.1636}& \textbf{+0.1683} & \textbf{+0.1613}\\
\hline\noalign{\smallskip}
\end{tabular}
\end{center}
\end{table}

\begin{figure}
\centering
\setlength{\abovecaptionskip}{-0.05cm} 
\setlength{\belowcaptionskip}{-0.2cm} 
\includegraphics[width=\columnwidth]{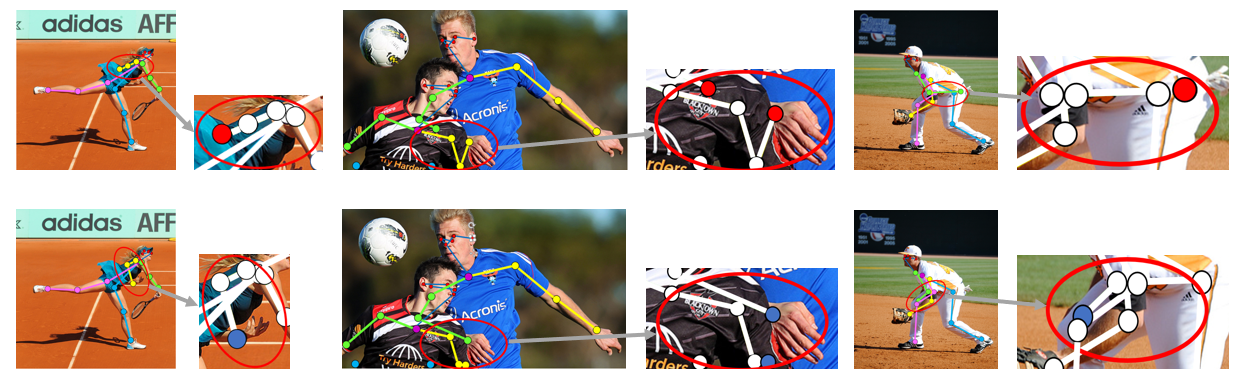}
\centering
\caption{Three examples of refinement of predicted keypoints. The top row is the original estimation. The bottom row is the refined version.}
\label{fig:example}
\end{figure}

\begin{figure}
\centering
\setlength{\abovecaptionskip}{-0.05cm} 
\setlength{\belowcaptionskip}{-0.4cm} 
\includegraphics[width=1\columnwidth]{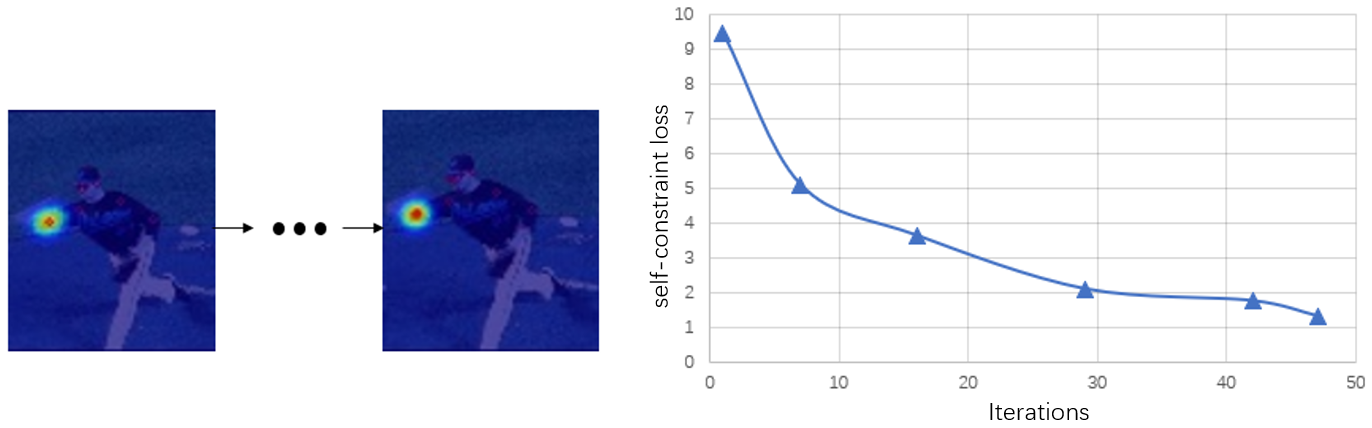}
\centering
\caption{The decreasing of the self-constraint loss during local search and refinement of the predicted keypoint.}
\label{fig:loss}
\end{figure}

\subsection{Ablation Studies}
To systematically evaluate our method and study the contribution of each algorithm component, we use the HRNet-W48 backbone to perform a number of ablation experiments on the COCO val2017 dataset. Our algorithm has two major new components, the Self-Constrained Learning (SCL) and the Self-Constrained  optimization (SCO). In the first row of Table \ref{tab:ablations}, we report the baseline (HRNet-W48) results. The second row shows the results with the SCL. The third row shows results with the SCL and SCO of the prediction results. We can clearly see that each algorithm component is contributing significantly to the overall performance. In Table \ref{tab:ablations of keypoints}, We also use normalization and sigmoid functions to evaluate the loss of terminal keypoints, and the results show that the confidence of each keypoint from HRNet has been greatly improved after using SCIO.

Fig. \ref{fig:example} shows three examples of how the estimation keypoints have been refined by the self-constrained inference optimization method. The top row shows the original estimation of the keypoints. The bottom row shows the refined estimation of the keypoints. Besides each result image, we show the enlarged image of those keypoints whose estimation errors are large in the original method. However, using our self-constrained optimization method, these errors have been successfully corrected.
Fig. \ref{fig:loss} shows how the self-constraint loss decreases during the search process. We can see that the loss drops quickly and the keypoints have been refined to the correct locations. 
In the Supplemental Materials, we provide additional experiments and algorithm details for further understanding of the proposed SCIO method.


\section{Conclusion}
In this work, we observed that  human poses exhibit strong structural correlation  within keypoint groups, which can be  explored to improve the accuracy and robustness of their estimation. 
We developed a self-constrained prediction-verification network to learn this coherent spatial structure and to perform local refinement of the pose estimation results during the inference stage. 
We partition each keypoint group into two subsets, base keypoints and terminal keypoints, and develop a self-constrained prediction-verification network  to perform forward and backward predictions between them. This prediction-verification network design is able  to capture the local structural correlation between keypoints. 
Once successfully learned, we used the verification network as a feedback module to guide the local optimization of pose estimation results for low-confidence keypoints with the self-constraint loss on high-confidence keypoints as the objective function. Our extensive experimental results on benchmark MS COCO datasets demonstrated that the proposed SCIO method is able to significantly improve the pose estimation results. 

 
%
%
\bibliographystyle{splncs04}
\bibliography{egbib}

\begin{thebibliography}{10}
\providecommand{\url}[1]{\texttt{#1}}
\providecommand{\urlprefix}{URL }
\providecommand{\doi}[1]{https://doi.org/#1}

\bibitem{DBLP:conf/cvpr/BagautdinovAFFS17}
Bagautdinov, T.M., Alahi, A., Fleuret, F., Fua, P., Savarese, S.: Social scene
  understanding: End-to-end multi-person action localization and collective
  activity recognition. In: {CVPR}. pp. 3425--3434 (2017)

\bibitem{DBLP:conf/cvpr/CaoSWS17}
Cao, Z., Simon, T., Wei, S., Sheikh, Y.: Realtime multi-person 2d pose
  estimation using part affinity fields. In: CVPR. pp. 1302--1310 (2017)

\bibitem{Chen_2018_CVPR}
Chen, Y., Wang, Z., Peng, Y., Zhang, Z., Yu, G., Sun, J.: Cascaded pyramid
  network for multi-person pose estimation. In: CVPR. pp. 7103--7112 (2018)

\bibitem{Cheng_2020_CVPR}
Cheng, B., Xiao, B., Wang, J., Shi, H., Huang, T.S., Zhang, L.: Higherhrnet:
  Scale-aware representation learning for bottom-up human pose estimation. In:
  CVPR. pp. 5385--5394 (2020)

\bibitem{DBLP:conf/cvpr/ElhayekAJTPABST15}
Elhayek, A., de~Aguiar, E., Jain, A., Tompson, J., Pishchulin, L., Andriluka,
  M., Bregler, C., Schiele, B., Theobalt, C.: Efficient convnet-based
  marker-less motion capture in general scenes with a low number of cameras.
  In: {CVPR}. pp. 3810--3818 (2015)

\bibitem{Fang_2017_ICCV}
Fang, H.S., Xie, S., Tai, Y.W., Lu, C.: Rmpe: Regional multi-person pose
  estimation. In: ICCV. pp. 2353--2362 (2017)

\bibitem{8575519}
Fieraru, M., Khoreva, A., Pishchulin, L., Schiele, B.: Learning to refine human
  pose estimation. In: 2018 IEEE/CVF Conference on Computer Vision and Pattern
  Recognition Workshops (CVPRW). pp. 318--31809 (2018)

\bibitem{Geng_2021_CVPR}
Geng, Z., Sun, K., Xiao, B., Zhang, Z., Wang, J.: Bottom-up human pose
  estimation via disentangled keypoint regression. In: CVPR. pp. 14676--14686
  (2021)

\bibitem{DBLP:conf/avss/GoldaKSB19}
Golda, T., Kalb, T., Schumann, A., Beyerer, J.: Human pose estimation for
  real-world crowded scenarios. In: {AVSS}. pp.~1--8 (2019)

\bibitem{He_2017_ICCV}
He, K., Gkioxari, G., Dollar, P., Girshick, R.: Mask r-cnn. In: ICCV. pp.
  2980--2988 (2017)

\bibitem{DBLP:conf/cvpr/0005ZGH20}
Huang, J., Zhu, Z., Guo, F., Huang, G.: The devil is in the details: Delving
  into unbiased data processing for human pose estimation. In: {CVPR}. pp.
  5699--5708 (2020)

\bibitem{DBLP:conf/iccv/HuangGT17}
Huang, S., Gong, M., Tao, D.: A coarse-fine network for keypoint localization.
  In: {ICCV}. pp. 3047--3056 (2017)

\bibitem{9107502}
Kamel, A., Sheng, B., Li, P., Kim, J., Feng, D.D.: Hybrid refinement-correction
  heatmaps for human pose estimation. IEEE Transactions on Multimedia
  \textbf{23},  1330--1342 (2021). \doi{10.1109/TMM.2020.2999181}

\bibitem{Khirodkar_2021_ICCV}
Khirodkar, R., Chari, V., Agrawal, A., Tyagi, A.: Multi-instance pose networks:
  Rethinking top-down pose estimation. In: Proceedings of the IEEE/CVF
  International Conference on Computer Vision (ICCV). pp. 3122--3131 (October
  2021)

\bibitem{DBLP:conf/cvpr/LiWZMFL19}
Li, J., Wang, C., Zhu, H., Mao, Y., Fang, H., Lu, C.: Crowdpose: Efficient
  crowded scenes pose estimation and a new benchmark. In: {CVPR}. pp.
  10863--10872 (2019)

\bibitem{DBLP:journals/corr/abs-1901-00148}
Li, W., Wang, Z., Yin, B., Peng, Q., Du, Y., Xiao, T., Yu, G., Lu, H., Wei, Y.,
  Sun, J.: Rethinking on multi-stage networks for human pose estimation. CoRR
  \textbf{abs/1901.00148} (2019)

\bibitem{DBLP:conf/eccv/LinMBHPRDZ14}
Lin, T., Maire, M., Belongie, S.J., Hays, J., Perona, P., Ramanan, D.,
  Doll{\'{a}}r, P., Zitnick, C.L.: Microsoft {COCO:} common objects in context.
  In: ECCV. pp. 740--755 (2014)

\bibitem{liu2021watching}
Liu, X., Zhang, P., Yu, C., Lu, H., Yang, X.: Watching you: Global-guided
  reciprocal learning for video-based person re-identification. In: Proceedings
  of the IEEE/CVF Conference on Computer Vision and Pattern Recognition. pp.
  13334--13343 (2021)

\bibitem{long2015fully}
Long, J., Shelhamer, E., Darrell, T.: Fully convolutional networks for semantic
  segmentation. In: Proceedings of the IEEE conference on computer vision and
  pattern recognition. pp. 3431--3440 (2015)

\bibitem{Luo_2021_CVPR}
Luo, Z., Wang, Z., Huang, Y., Wang, L., Tan, T., Zhou, E.: Rethinking the
  heatmap regression for bottom-up human pose estimation. In: CVPR. pp.
  13264--13273 (2021)

\bibitem{DBLP:conf/cvpr/MoonCL19}
Moon, G., Chang, J.Y., Lee, K.M.: Posefix: Model-agnostic general human pose
  refinement network. In: {CVPR}. pp. 7773--7781 (2019)

\bibitem{DBLP:conf/nips/NewellHD17}
Newell, A., Huang, Z., Deng, J.: Associative embedding: End-to-end learning for
  joint detection and grouping. In: {NeurIPS}. pp. 2277--2287 (2017)

\bibitem{DBLP:conf/cvpr/PapandreouZKTTB17}
Papandreou, G., Zhu, T., Kanazawa, N., Toshev, A., Tompson, J., Bregler, C.,
  Murphy, K.: Towards accurate multi-person pose estimation in the wild. In:
  CVPR. pp. 3711--3719 (2017)

\bibitem{DBLP:conf/cvpr/RhodinCKSF19}
Rhodin, H., Constantin, V., Katircioglu, I., Salzmann, M., Fua, P.: Neural
  scene decomposition for multi-person motion capture. In: {CVPR}. pp.
  7703--7713 (2019)

\bibitem{DBLP:conf/cvpr/SuYXGW19}
Su, K., Yu, D., Xu, Z., Geng, X., Wang, C.: Multi-person pose estimation with
  enhanced channel-wise and spatial information. In: {CVPR}. pp. 5674--5682.
  Computer Vision Foundation / {IEEE} (2019)

\bibitem{Sun_2020_CVPR}
Sun, H., Zhao, Z., He, Z.: Reciprocal learning networks for human trajectory
  prediction. In: CVPR. pp. 7414--7423 (2020)

\bibitem{DBLP:conf/cvpr/0009XLW19}
Sun, K., Xiao, B., Liu, D., Wang, J.: Deep high-resolution representation
  learning for human pose estimation. In: {CVPR}. pp. 5693--5703 (2019)

\bibitem{DBLP:conf/eccv/SunXWLW18}
Sun, X., Xiao, B., Wei, F., Liang, S., Wei, Y.: Integral human pose regression.
  In: {ECCV}. pp. 536--553 (2018)

\bibitem{DBLP:conf/eccv/WangLGDW20}
Wang, J., Long, X., Gao, Y., Ding, E., Wen, S.: Graph-pcnn: Two stage human
  pose estimation with graph pose refinement. In: ECCV. pp. 492--508 (2020)

\bibitem{DBLP:conf/cvpr/WangTM20}
Wang, M., Tighe, J., Modolo, D.: Combining detection and tracking for human
  pose estimation in videos. In: {CVPR}. pp. 11085--11093 (2020)

\bibitem{DBLP:conf/cvpr/WuWWGW19}
Wu, J., Wang, L., Wang, L., Guo, J., Wu, G.: Learning actor relation graphs for
  group activity recognition. In: {CVPR}. pp. 9964--9974 (2019)

\bibitem{DBLP:conf/eccv/XiaoWW18}
Xiao, B., Wu, H., Wei, Y.: Simple baselines for human pose estimation and
  tracking. In: ECCV. pp. 472--487 (2018)

\bibitem{xu2020segmentation}
Xu, C., Howey, J., Ohorodnyk, P., Roth, M., Zhang, H., Li, S.: Segmentation and
  quantification of infarction without contrast agents via spatiotemporal
  generative adversarial learning. Medical image analysis  \textbf{59},  101568
  (2020)

\bibitem{DBLP:conf/cvpr/XuT21}
Xu, T., Takano, W.: Graph stacked hourglass networks for 3d human pose
  estimation. In: CVPR. pp. 16105--16114 (2021)

\bibitem{DBLP:conf/cvpr/YangRLZW021}
Yang, Y., Ren, Z., Li, H., Zhou, C., Wang, X., Hua, G.: Learning dynamics via
  graph neural networks for human pose estimation and tracking. In: {CVPR}. pp.
  8074--8084 (2021)

\bibitem{DBLP:journals/corr/abs-1905-05355}
Yu, D., Su, K., Geng, X., Wang, C.: A context-and-spatial aware network for
  multi-person pose estimation. CoRR  \textbf{abs/1905.05355} (2019)

\bibitem{DBLP:conf/cvpr/ZhangZD0Z20}
Zhang, F., Zhu, X., Dai, H., Ye, M., Zhu, C.: Distribution-aware coordinate
  representation for human pose estimation. In: {CVPR}. pp. 7091--7100 (2020)

\bibitem{zhang2021accurate}
Zhang, L., Zhou, S., Guan, J., Zhang, J.: Accurate few-shot object detection
  with support-query mutual guidance and hybrid loss. In: Proceedings of the
  IEEE/CVF Conference on Computer Vision and Pattern Recognition. pp.
  14424--14432 (2021)

\bibitem{Zhu_2017_ICCV}
Zhu, J.Y., Park, T., Isola, P., Efros, A.A.: Unpaired image-to-image
  translation using cycle-consistent adversarial networks. In: ICCV. pp.
  2242--2251 (2017)

\end{thebibliography}
\end{document}